\pdfoutput=1

\documentclass[11pt]{article}

\usepackage[]{acl}

\usepackage{times}
\usepackage{latexsym}
\usepackage{amssymb}

\usepackage[T1]{fontenc}

\usepackage[utf8]{inputenc}

\usepackage{microtype}
\usepackage{hyperref}

\usepackage{inconsolata}

\usepackage{graphicx}

\usepackage{longtable}
\usepackage{afterpage}
\usepackage{hyperref}
\usepackage{makecell}
\usepackage{xcolor}

\title{Can NLP Tackle Hate Speech in the Real World?\\ Stakeholder-Informed Feedback and Survey on Counterspeech}

\author{Tanvi Dinkar,$^{1}$   Aiqi Jiang,$^{1}$ Simona Frenda,$^{1}$ Poppy Gerrard-Abbott,$^{2}$\\ \textbf{ Nancie Gunson},$^{1}$  
        \textbf{Gavin Abercrombie}$^{1}$ \and \textbf{Ioannis Konstas}$^{1}$\\
        $^{1}$The Interaction Lab, Heriot-Watt University, $^{2}$University of Edinburgh \\ 
        \texttt{\{t.dinkar, a.jiang, s.frenda, n.gunson, g.abercrombie, i.konstas\}@hw.ac.uk}\\
        \texttt{pgerrard@exseed.ed.ac.uk}}

\begin{document}
\maketitle
\begin{abstract}
Counterspeech, i.e. the practice of responding to online hate speech, has gained traction in NLP as a promising intervention. 
While early work emphasised collaboration with non-governmental organisation stakeholders, recent research trends have shifted toward automated pipelines that reuse a small set of legacy datasets, often without input from affected communities. This paper presents 
a systematic review of 74 NLP studies on counterspeech, analysing the extent to which stakeholder participation influences dataset creation, model development, and evaluation. To complement this analysis, we conducted a participatory case study with five NGOs specialising in online Gender-Based Violence (oGBV), identifying stakeholder-informed practices for counterspeech generation. Our findings reveal a growing disconnect between current NLP research and the needs of communities most impacted by toxic online content. We conclude with concrete recommendations for re-centring stakeholder expertise in counterspeech research.\looseness=-1

\end{abstract}

\section{Introduction}

The automation of counterspeech responses to toxic online content such as hate speech and disinformation is a growing topic in Natural Language Processing (NLP)~\citep{bonaldi-etal-2024-nlpw}.
At the same time, there has been increasing recognition that NLP research should aim to focus on the needs of the stakeholders that the tools it develops are designed to serve community (i.e. through participatory design)~\citep{birhane-etal-2022-power,caselli-etal-2021-guiding}, particularly when it comes to tackling hate speech~\citep{abercrombie-etal-2023-resources,parker-ruths-2023-hate}.\looseness=-1

Inspired by the work of non-governmental organisations (NGOs) engaged in toxicity countering,\footnote{e.g. \url{https://getthetrollsout.org}} 
efforts at automating counterspeech generation began quite promisingly in this regard, with a focus on integrating experts at combating real-world online toxicity into human-in-the-loop systems in the \textsc{CONAN}\footnote{\url{https://github.com/marcoguerini/CONAN}}
family of datasets~\citep{bonaldi-etal-2022-human,chung-etal-2019-conan,fanton-etal-2021-human}.
However, as we show in this review, recent work has relied on automated research pipelines in which a few, now relatively old counterspeech datasets are repeatedly reworked with further layers of automatic and/or non-expert produced data, and stakeholders (outwith the computer scientists conducting the research) are typically not involved in their conception, development, or evaluation.

Where recent reviews of counterspeech research have focused on either synthesising findings from real-world counterspeech campaigns~\citep{chung-etal-2024-understanding} or technical aspects of natural language generation~\citep{bonaldi-etal-2024-nlpw}, we focus on stakeholder participation in NLP research in this work.

\paragraph{Our contributions}
We conduct a \textbf{systematic review} (§\ref{sec:review}) of 74 relevant publications focused on data resources, models, and computational analysis of counterspeech, and answer research question 1 (\textbf{RQ1}):
To what extent are affected stakeholders represented in NLP counterspeech research?

In analysing the results, 
we assess the reviewed work against 
insights from stakeholders and experts on the best approaches to counterspeech.
As a \textbf{case study} (§\ref{sec:case}), we discuss
findings from participatory design work with five NGOs in relation to our survey findings that work to tackle online Gender-Based Violence (oGBV), and investigate research question 2 (\textbf{RQ2}): 
What stakeholder-informed feedback practices can be used to counter hate?\looseness=-1 

Findings suggest that NLP research on counterspeech should be redirected towards the needs of such stakeholders. 
Based on the feedback and issues raised, we provide specific recommendations for NLP practitioners to produce stakeholder-informed counterspeech (§\ref{sec:recommendations}).

\section{
Background and key concepts}

As an alternative to content removal,
\textbf{Counterspeech}
refers to responses that challenge toxic online content, 
and is seen as a promising way of tackling hate.
In NLP, research  has focused on creating datasets \citep{Mathew2018ThouSN,Chung2021TowardsKC}, 
developing automated counterspeech generation systems \citep{bonaldi-etal-2023-weigh,gupta-etal-2023-counterspeeches}, and designing (usually intrinsic) evaluation methods \citep{Zubiaga2024ALR,Halim2023WokeGPTIC}. 
In sociology, \citet{buerger2019counterspeech} and \citet{alsagheer2022counter} review 
recent trends 
in counterspeech and provide general introductions to its concept, features and applications, while \citet{benesch2016counterspeech} propose a taxonomy of strategies used to counter hate online. 
From an NLP perspective, \citet{chung-etal-2024-understanding} survey the dynamics and effectiveness of counterspeech, and \citet{bonaldi-etal-2024-nlpw} the methods and challenges involved in its automation. 
\citet{tomalin-ullman-2023-counterspeech} 
contribute by compiling multidisciplinary perspectives on counterspeech, including its automation and evaluation. This survey addresses existing gaps by highlighting the importance of stakeholder perspectives in developing counterspeech.

The growing application of AI systems for social good~\cite{moorosi2023ai} has increased the engagement of stakeholders in research; with different structures, principles and modalities to guide \textbf{participatory design}
\cite{caselli-etal-2021-guiding,birhane-etal-2022-power,delgado2023participatory}. 
However, \citet{parker-ruths-2023-hate} have identified a disconnect between computer science research and affected communities when it comes to tackling
hate speech and its consequences. 
They propose key points to create a more integrated community to address this: involving groups that combat hate speech who have a deeper understanding of responses to hate speech and its impact on society. In this context, participatory design, popular in  branches of computer science such as human-computer interaction \citep{muller1993participatory},  
gives a voice in the design process to people who lack expert design skills.

Whilst not explicitly referencing participatory methodologies, several early NLP works on counterspeech engaged with domain expert stakeholders to create human-in-the-loop generation pipelines~\citep{chung-etal-2019-conan,bonaldi-etal-2022-human,Fanton2021HumanintheLoopFD}. More recently, \citet{mun-etal-2024-counterspeakers} conducted a large-scale survey with relevant stakeholders to inform the design of NLP counterspeech tools~\citep{mun-etal-2024-counterspeakers}.
In this review, we 
uncover 
the extent to which stakeholders participate in NLP counterspeech research design and resource creation.

\begin{figure}
    \centering
    \includegraphics[width=0.8\columnwidth]{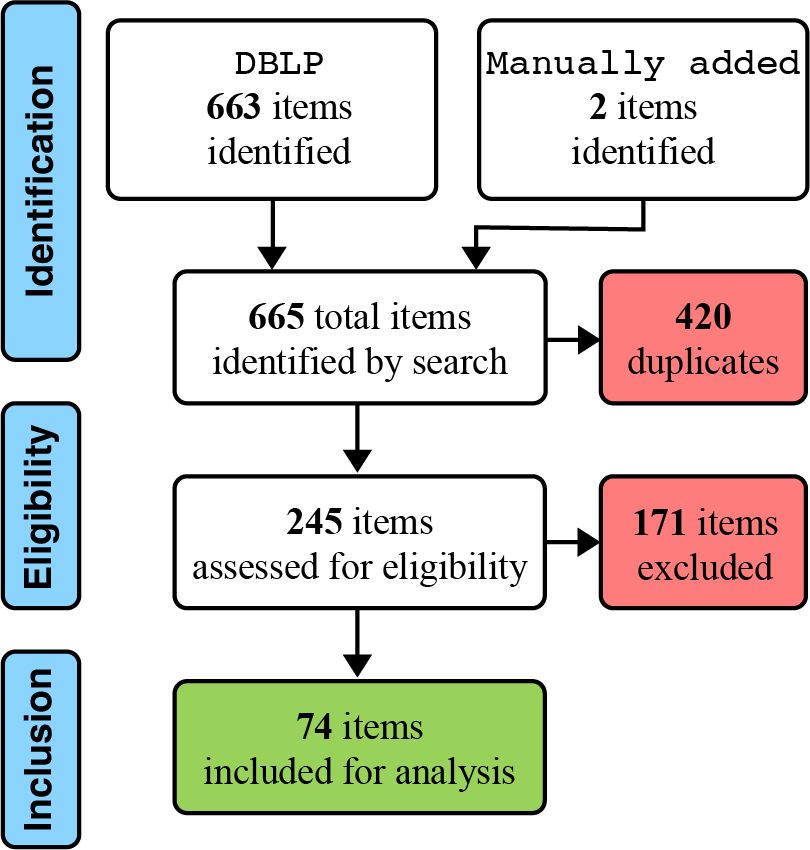}
    \caption{Search and selection protocol.}
    \label{fig:protocol}
\end{figure}

\textbf{Online Gender-Based Violence} or \emph{oGBV} is a framework used by international organisations such as the UN and WHO, and covers harmful effects on all genders, 
particularly women.\footnote{\url{https://www.who.int/health-topics/violence-against-women}} Misogynistic abuse affects around 50\% of women and especially further marginalised groups~\citep{glitch-ewaw-2020-ripple,parikh-etal-2019-multi}, resulting in women often feeling uncomfortable online \citep{stevens2024women}. Although 
there have been recent efforts to identify oGBV, 
including various SEMEVAL tasks ~\citep{basile-etal-2019-semeval,fersini-etal-2022-semeval,kirk-etal-2023-semeval}, existing computational approaches and datasets suffer from several shortcomings ~\citep{abercrombie-etal-2023-resources}, such as the lack of participation in designing taxonomies and formalisms of the addressed social problem, and the exclusion, due to the adopted terminology, of specific aspects related to various forms of violence. 
As a counterspeech case study,
we describe the experiences of 
expert stakeholders in addressing oGBV, carrying out focus groups that involved victims/survivors, bystanders and professional supporters of victims.

\section{Systematic Review}
\label{sec:review}

\begin{table*}[t]
\centering
\fontsize{6.5}{7.6}\selectfont

\label{tab:source_datasets}
\begin{tabular}{p{2.5cm}p{2.9cm}p{3.5cm}p{2.2cm}p{2.5cm}}
\textbf{Publication} & \textbf{HS source} & \textbf{CS source} & \textbf{Human input and Task (None = $\times$)} & \textbf{Stakeholder involvement (\checkmark/$\times$) with Details} \\
\hline
\begin{tabular}[t]{@{}l@{}}$\heartsuit$ \textsc{CONAN} \\\citep{chung-etal-2019-conan}\end{tabular}  & Nichesourcing & Nichesourcing & \begin{tabular}[t]{@{}l@{}}Write HS/CS +\\ Paraphrase CS\end{tabular} & \begin{tabular}[t]{@{}l@{}}$\checkmark$NGO workers, $\times$non-experts\end{tabular} \\
\begin{tabular}[t]{@{}l@{}}$\spadesuit$ \textsc{Multi-target CONAN} \\ \citep{Fanton2021HumanintheLoopFD}\end{tabular} & \begin{tabular}[t]{@{}l@{}}Hybrid: Nichesourcing and \\Automated (Human-in-the-loop)\end{tabular} & \begin{tabular}[t]{@{}l@{}}Hybrid: Nichesourcing and \\Automated (Human-in-the-loop)\end{tabular} & \begin{tabular}[t]{@{}l@{}}Val CS + Edit CS\end{tabular} & \begin{tabular}[t]{@{}l@{}}$\checkmark$NGO workers, $\times$academics\end{tabular} \\
\begin{tabular}[t]{@{}l@{}}$\clubsuit$ \textsc{DIALOCONAN} \\ \citep{bonaldi-etal-2022-human}\end{tabular} & \begin{tabular}[t]{@{}l@{}}Hybrid: Nichesourcing and \\Automated (Human-in-the-loop)\end{tabular} & \begin{tabular}[t]{@{}l@{}} Hybrid: Nichesourcing and \\Automated (Human-in-the-loop)\end{tabular} &  \begin{tabular}[t]{@{}l@{}}Val CS + Edit CS\end{tabular} & $\checkmark$NGO workers \\
\begin{tabular}[t]{@{}l@{}}$\Box$ \textsc{MTKGCONAN}\\ \citep{Chung2021TowardsKC}\end{tabular} & Existing dataset ($\heartsuit$) & Automated generation & \begin{tabular}[t]{@{}l@{}}Ann/Eval CS\end{tabular} &  $\checkmark$NGO workers \\
\begin{tabular}[t]{@{}l@{}}\textsc{IntentCONAN} \\ \citep{gupta-etal-2023-counterspeeches}\end{tabular} & Existing dataset ($\spadesuit$) & \begin{tabular}[t]{@{}l@{}} Existing dataset ($\spadesuit$) + \\Human written\end{tabular} & Write CS & $\times$academics \\
\begin{tabular}[t]{@{}l@{}}\textsc{ML-MTCONAN-KN}\\ \citep{bonaldi-etal-2025-first}\end{tabular} & Existing dataset ($\Box$) & Human written & \begin{tabular}[t]{@{}l@{}}Write CS + \\Edit MT HS/CS\end{tabular} & $\times$academics: translators Spanish, Basque, Italian\\
\begin{tabular}[t]{@{}l@{}}$\Diamond$ \textsc{Benchmark}\\ \citep{qian-etal-2019-benchmark}\end{tabular} & \begin{tabular}[t]{@{}l@{}}Hybrid: Crawling + \\Crowdsourcing\end{tabular} & \begin{tabular}[t]{@{}l@{}}Crowdsourcing + \\Automated generation\end{tabular} & Val HS + Write CS & $\times$crowdworkers\\
\hline
\end{tabular}
\caption{Summary of frequently used existing datasets in counterspeech. The table reports hate speech (HS) and counterspeech (CS) data sources, the type of human input involved in any research stages (`Val': Validating HS/CS instances, `Ann/Eval': Annotate/Evaluate), and the extent of stakeholder involvement.  We list datasets that are used more than twice for both HS and CS sources across the surveyed resources, but exclude those used more than twice for only HS. Note, the `Hybrid' label is only used when different methods are used within one HS or CS instance; using automated methods to generate CS and then nichesourcing to correct the same CS. 
The bracket in the last column gives details about the human involvement within the symbol (\checkmark/$\times$), row 1 shows that (NGO workers) are stakeholders given \checkmark, with non-experts written as outside the bracket.}
\label{tab:source_datasets}
\end{table*}

We conducted a systematic review of computer science publications on the topic of counterspeech, following the PRISMA methodology~\citep{moher-etal-2009-preferred}.
The review protocol is shown in \autoref{fig:protocol}. 

\begin{table}[!h]
    \centering
    \small
    \begin{tabular}{p{3.55cm}|p{3.55cm}}
        \textbf{Include} & \textbf{Exclude} \\
        \hline
        Resources related to human-written counterspeech for dataset creation. & Resources that contain the keyword `\emph{counter-terrorism}' in isolation with none of our other keywords. \\
        \hline
        Resources related to in-the-wild human-written counterspeech for social media analysis. & Resources with tasks that were irrelevant to the present work, such as \emph{speech-spoofing}.\\
        \hline
        Resources that do automated counterspeech generation. & Survey resources on counterspeech. \\ 
    \end{tabular}
    \normalsize 
    \caption{Inclusion/exclusion criteria for the review.}
    \label{tab:eligibility}
\end{table}

\paragraph{Identification}
To isolate relevant counterspeech research and exclude work from fields such as social science that are not concerned with NLP methods, we searched the computer science bibliography database \href{https://dblp.org/}{DBLP}. All searches were conducted in March 2025. Following \citet{chung-etal-2024-understanding}, we used the keywords `\emph{counter-speech}', `\emph{counter-
narratives}', `\emph{counter-terrorism}', `\emph{counter-aggression}',
`\emph{counter-hate}', `\emph{counter speech}', `\emph{counter narrative}',
`\emph{countering online hate speech}', `\emph{counter hate speech}',
and `\emph{counter-hate speech}', and additionally added the keyword `\emph{counterspeech}'.

\paragraph{Eligibility criteria} 
Overall, our goal is to focus on human-written and synthetically generated counterspeech resources in computer science, to answer questions regarding the ways the counterspeech data is sourced, and additionally the level of participatory design involved. 
\autoref{tab:eligibility} describes the inclusion and exclusion criteria that were applied. Using these criteria, two of the authors excluded and identified items to review, which were cross-checked by a third author. 
We then turned our attention to counterspeech resources based on `in-the-wild' data or performing social media analyses, as these resources may include opinions from experienced users in responding to hate speech online. 

\paragraph{Summary of included resources}
After following the systematic survey process, we were left with 74 items for systematic review that cover wholly or partially automatically generated counterspeech, and the computational analysis of real counterspeech in online settings. 

\subsection{Results and Discussion}
\label{sec: RQ1}

\begin{figure*}[ht!]
    \centering
    \includegraphics[width=1\textwidth]{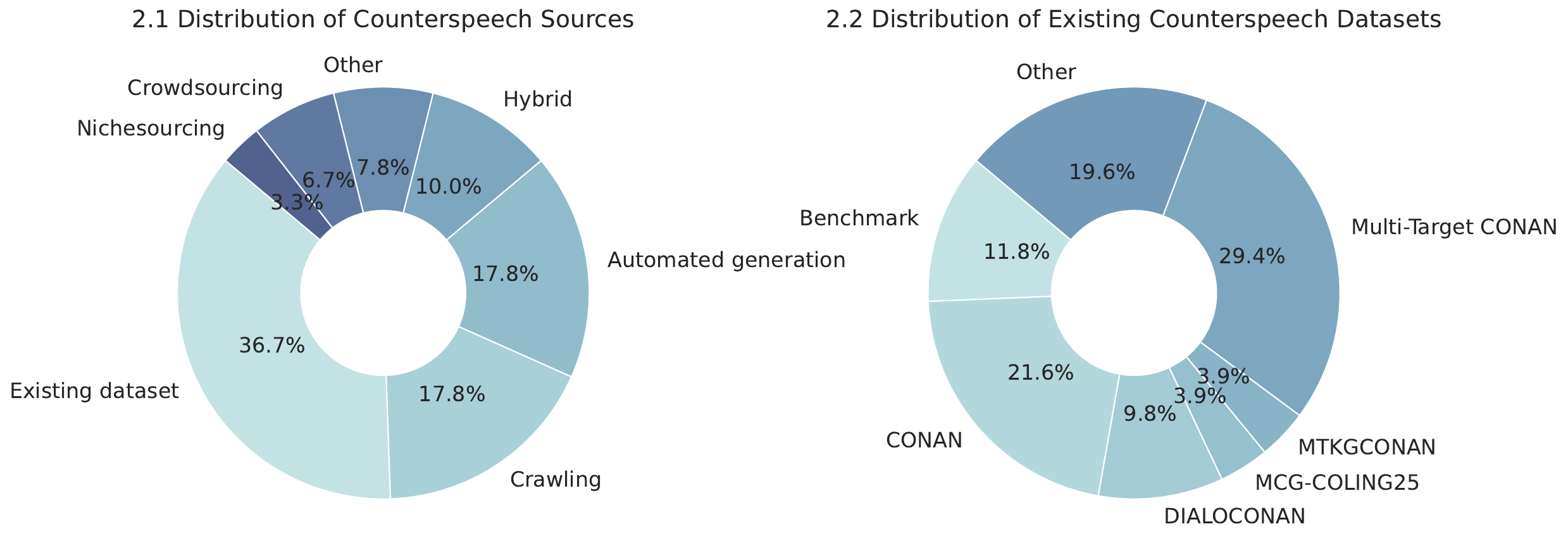}
    \caption{Counterspeech sources and datasets. The percentage reflects the proportion of total sources (N = 88), given that some resources include more than one source.}
    \label{fig:cs_source}
\end{figure*}

\paragraph{Preliminary findings.} The results of our survey are given in \autoref{tab:source_datasets}, which outlines the most commonly used datasets in counterspeech research and \autoref{tab:datasets}, which consists of the rest of the surveyed resources. As visually shown in \autoref{tab:datasets}, close to $50\%$ of the surveyed resources use an existing dataset for sourcing hate speech or counterspeech\footnote{Indeed, it was difficult to initially identify whether different resources used the same dataset, given different naming conventions to refer to the same dataset.}. Of these resources, as shown in \autoref{fig:cs_source} (right), $66\%$ use an iteration of the CONAN \citep{chung-etal-2019-conan} dataset, i.e. Multi-Target CONAN \citep{Fanton2021HumanintheLoopFD}, DIALOCONAN \citep{bonaldi-etal-2022-human} or MTKGCONAN \citep{Chung2021TowardsKC}. This is concerning, as constant re-use of these datasets (indeed without benchmarks for comparison and difficulties formulating metrics that capture high-quality counterspeech) can lead to a ceiling effect in terms of performance. 
Additionally, the majority of the source datasets were created before LLMs were widely adopted (e.g. CONAN in 2019, Multi-Target CONAN in 2021); these datasets may have been used in the training of proprietary or closed-source models \citep{balloccu-etal-2024-leak}, making it difficult to assess such models fairly for automated counterspeech generation (memorising exact responses to the hate speech, or source datasets containing outdated examples of hate speech)\footnote{However, this is currently speculative and warrants further research.}. \autoref{fig:cs_source} (left) also shows that Nichesourcing or relying on experts to produce counterspeech \citep{bonaldi-etal-2024-nlpw}, is the least used method to source counterspeech. We additionally analysed the sources for the modes of participatory design according to \citet{delgado2023participatory}, to mark 6 of the resources as \emph{`Consult'}, with an additional 4 as \emph{`Consult/Include(?)'}.

\begin{figure}[ht!]
    \centering
    \includegraphics[width=1.1\columnwidth]{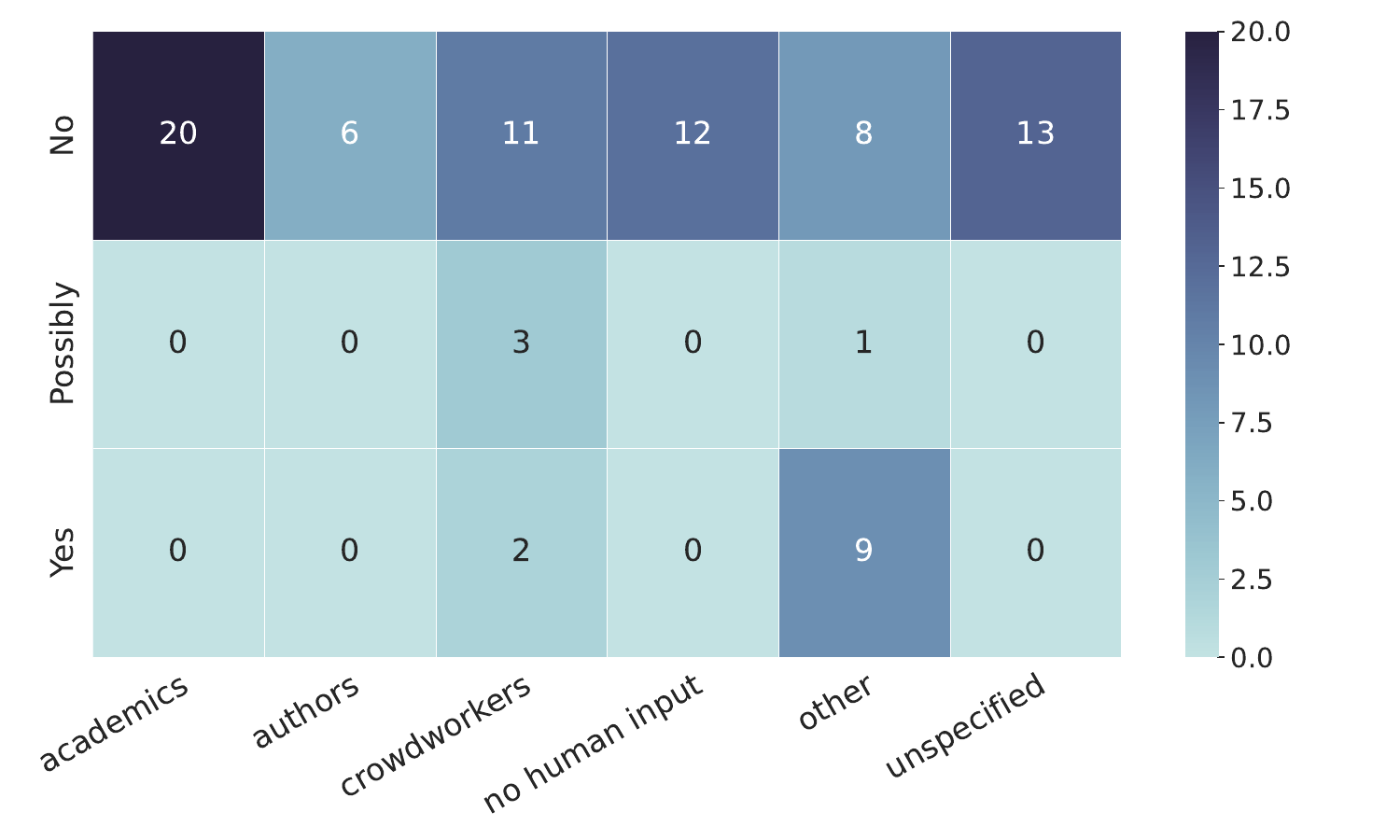}
    \caption{Stakeholder identity by participation. 
    `Possibly' indicates bystander participation.}
    \label{fig:stakeholder}
\end{figure}

\paragraph{What is an expert and the value of `non-expertise'} Our survey indicates that counterspeech resources use the word \emph{`expert'} in two different ways. \citet{chung-etal-2019-conan,Tekiroglu2020GeneratingCN, Chung2021TowardsKC, bonaldi-etal-2022-human, Chung2024OnTE, Jones2024AMF} use this term specifically to distinguish NGO workers from non-expert crowdworkers. We also see use of the word \emph{`expert'} when a professional/expert translator is engaged, \citet{Chung2020ItalianCN} for Italian, or \citet{Bengoetxea2024BasqueAS} for Spanish and Basque, and \citet{bonaldi-etal-2025-first} for all three. However, another group of resources uses this term to indicate domain knowledge in computer science, NLP or linguistics such as in \citet{gupta-etal-2023-counterspeeches,Mun2023BeyondDH, Saha2024CrowdCounterAB,Hengle2024IntentconditionedAN}, possibly to distinguish this from data collected from crowdworkers. In  \autoref{tab:datasets}, the latter group can be seen in the column \emph{`Stakeholder involvement'} where we have distinguished between whether the \emph{`experts'} are the authors themselves or other academics with pan-NLP domain expertise. We also use the \emph{`academic'} label when resources don't necessarily claim expert involvement, but do specify academic qualifications as the criteria for annotator recruitment (`3 grad students'). As \autoref{fig:stakeholder} shows, $26$ of the resources we surveyed use either the authors themselves or other academics to annotate or evaluate counterspeech.

Regarding non-experts, some resources may deliberately use crowdworkers to annotate/evaluate counterspeech, such as in \citet{Jones2024AMF}, to get opinions on how difficult it is for an everyday social media user to write counterspeech based on \emph{expert-written NGO guidelines}, and what the barriers are that prevent them from doing so.

\paragraph{Stakeholder and bystander participation}
It is important to define the terms `stakeholder' and `bystander' in order to explain our labelling process in the \emph{`Stakeholder involvement'} column in \autoref{tab:datasets}. \emph{Stakeholders} refer to agents who practice a niche \emph{`stake'} in interests and processes, such as civil or campaigning gains \textit{[\dots] ``individuals, groups or organisations that share common interests and hold interest in the outcomes of certain decisions or objectives [\dots]''} \citep{chidwick2024contradictions}. Whilst traditionally referring to business, and often a contested term in feminist research \citep{Wicks_Gilbert_Freeman_1994}, the label is now understood to apply to a range of organisations \citep{miles2017stakeholder}, from policymaking to the third sector. \emph{Bystander} refers to a member of the public and/or community member (who is also a user if referring to internet spaces) who is a first-hand witness to hate speech and holds decision-making power around active and inactive responses, and is a secondary party involved in vicarious trauma.

In our survey, we expand on stakeholder participation to include bystander participation (as shown with the label \emph{`Possibly'} in \autoref{tab:datasets}). 
e.g. \citet{Lee2023AlternativeSC} recruited annotators with the \emph{explicit requirement} that the annotators have spent time online and encountered hate speech. \citet{Ping2024BehindTC,Ding2024CounterQuillIT} recruit participants across the US to research (a) why participants may be inclined/disinclined to participate in counterspeech writing online, (b) the frequency with which participants write counterspeech, and (c) participants' opinions on using AI tools to aid in counterspeech writing. While \citet{mun-etal-2024-counterspeakers} utilise both NGO workers and Amazon Mechanical Turk (AMT) workers, there is possible stakeholder participation from (only) the latter, as $94\%$ of the workers reported to have encountered hate speech online and $70\%$ had experience responding to the hate speech. These resources aim to understand more generalised opinions of bystanders on what are the barriers preventing people from engaging in counterspeech online
\footnote{Note, we did not use the \emph{`Possibly'} label for research that used professional translators to edit machine-translated hate speech/counterspeech pairs \citep{Bengoetxea2024BasqueAS}, native speakers that wrote low-resource Bengali and Hindi counterspeech \citep{Das2024LowResourceCG}, evaluated counterspeech in Italian \citep{Chung2020ItalianCN}, German \citep{Garland2020CounteringHO}, English, e.g. \citet{Zhu2021GeneratePS}, or annotators of Asian descent who annotated anti-Asian COVID-19 related hate speech/counterspeech tweets \citep{Ziems2020RacismIA} as details about their opinions and lived experiences regarding online hate speech and counterspeech are not provided.}.

\paragraph{Barriers to participatory design} 
(Lack of) funding and network can create huge barriers to participatory design. While this work focuses on the level of stakeholder involvement in counterspeech resources, we acknowledge these factors as challenges in having such involvement. 
One of the surveyed papers, \citet{Jones2024AMF}, explain their use of crowdworkers due to \textit{``[\dots] lack of direct access to expert NGO operators [\dots]''}. As outlined in \citet{caselli-etal-2021-guiding}, obtaining funding offers an additional barrier to participatory design research. 

\section{Case study: 
Addressing online Gender-Based Violence}
\label{sec:case}
While the practices followed by the CONAN datasets centred stakeholder participation, the results of our systematic survey show that this initial goal has been somewhat lost in the resources that followed. An increasing number of datasets reuse the same data with newer algorithmic methods. 
To understand whether there exist practices used in real world counterspeech that the NLP community is yet to adopt, we conducted a series of structured interactive focus groups~\citep{morgan-1996-focus} to get stakeholder input on  countering hate online, using feminist co-creation and participatory action design practices \citep{Askins02092018}. Our goal is to compile high-level feedback from stakeholders on countering hate online that will be relevant to the NLP community.

We invited oGBV organisations\footnote{Given our specific network of contacts, we decided to focus on the topic of oGBV.} on a country-wide basis. 
In each focus group, we asked for stakeholder input by deploying open-ended unstructured
questions about oGBV into collaborative practical activities \citep{Goessling2025}. 
This activity consisted of working with the stakeholders to identify real-world hate-speech samples we collected\footnote{These samples were manually collected mainly from X/Twitter and included both text and image examples.} and get their feedback on the best ways to respond. In the focus groups the authors adopted an observatory and note-taking role, while the stakeholders discussed their insights. At the start of the focus group, we included a high-level explanation of \emph{`AI'}-generated counterspeech, for stakeholders to understand the scope of our project from a computer science perspective. 
\autoref{tab:NGO_details} gives a brief description of these organisations. Each charity has different specialist focuses, leading to diverse perspectives on counterspeech approaches to oGBV.

\begin{table}[!h]
    \centering
    \small
    \begin{tabular}{p{2.15cm}|p{4.9cm}}
        \textbf{NGO} & \textbf{Areas of work and expertise} \\
        \hline
        A. \textsc{EVAW} \href{https://www.endviolenceagainstwomen.org.uk}{https:// www.endviolence againstwomen.org.uk} & A representative collective of violence against women organisations lobbying government for feminist policy on GBV. \\
        \hline
        B. \textsc{Glitch}  \href{https://glitchcharity.co.uk/}{https://glitch charity.co.uk/} & A national charity focused on oGBV especially
        towards Black women, producing best practice guidance and recommendations for tech companies and government.\\
        \hline
        C. \hspace{-0.25cm} \textsc{Amina} \ \ \ \href{https://mwrc.org.uk/}{https:// mwrc.org.uk} & A local charity focusing on empowering Muslim and Black \& Minority Ethnic (BME) women. Work includes running a helpline to support victims/survivors, providing legal advice regarding immigration concerns and campaigning.\\
        \hline
        D. \textsc{Scottish Women's Aid} \href{https://womensaid.scot/}{https:// womensaid.scot/} & A government-funded charity running advice services for domestic abuse victims. \textit{Note: The NGO worker who participated in this focus group was an expert in financial and online abuse.}\\
        \hline
        E. \textsc{Compass Centre} \href{https://www.compasscentre.org/}{https://www. compasscentre. org/} & A small rural GBV charity providing support and advocacy for rape and sexual violence victims/survivors, including a youth group and phone service. \textit{Note: Our focus group specifically engaged with people from the young persons' activist group within this NGO who were survivors of GBV}. 
    \end{tabular}
    \normalsize 
    \caption{Details of the NGOs that participated in focus groups to obtain expert insights on countering oGBV.}
    \label{tab:NGO_details}
\end{table}

\subsection{Results and Discussion}
\label{results_gbv}
In \autoref{sec:review}, we focused on results from our survey related to participatory design in existing counterspeech research; i.e., which datasets are used, the level and stage of human involvement, terminological discussions around the use of the word `\textit{expert}' to describe annotators, and stakeholder and bystander participation. In this section, we draw on our focus groups with NGOs to interpret and expand on additional survey findings. In particular, we focus on results from our survey that highlight missing elements in current research which would better align with stakeholder-informed feedback. Specifically, aspects of hate speech used to condition counterspeech (a prominent concern among the experts in our focus groups); i.e. missing metadata on the type of hate speech and its targets, lack of sub-categorisation of hate speech, and strategy use in NLP counterspeech. \underline{While these results are not discussed in \autoref{sec:review}}, we elaborate on them here to translate stakeholder feedback into concrete gaps we’ve identified through our survey. A summary of the feedback from the focus groups can be found in \autoref{tab:NGO_summary}.

Note: Early on, participants from \textsc{EVAW} used the terms \emph{perpetrator}, \emph{target} and \emph{bystander} to differentiate the roles involved in oGBV, which we adopt.\looseness=-1  

\begin{table}[!h]
    \centering
    \small
    \begin{tabular}{p{2.7cm}|p{4.4cm}}
        \textbf{Focus Issue} & \textbf{Reasoning} \\
        \hline
        \textit{Date of HS creation} & Interventions are time sensitive, replying to older content can bring further attention towards the HS. \\
        \hline
        \textit{Views and shares of HS} & Using these cues to determine if the HS warrants a reply (e.g. weighing benefits between intervening versus prioritising one's own safety). \\
        \hline
        \textit{Reach of the perpetrator} & Strategies to adopt differ depending on perpetrator reach.\\
        \hline
        \textit{Use of multiple strategies within the same counterspeech} & To answer to different parties involved, i.e. shutting down the perpetrator, providing resources for the target and educating bystanders. Note some of the NGOs had strict policies against engaging the perpetrator.\\
        \hline
        \textit{Sub-category of GBV} & Depending on sub-category of GBV (e.g. harassment versus dogpiling), different approaches are adopted. \\
        \hline
        \textit{Anthropomorphism of CS} & Weary of bots reinforcing stereotypical `feminazi' talking points, complications on bots that are explicitly gendered. \\
        \hline
        \textit{Temporality of Language} & Perpetrators engage in `algo-speak', finding new ways to escape being flagged by content moderation systems.  \\
    \end{tabular}
    \normalsize 
    \caption{Summary of key insights from NGOs.}
    \label{tab:NGO_summary}
\end{table}

\paragraph{A need for context.}
Perhaps the starkest difference between counterspeech-focused NLP and stakeholder input was the level of attention given to meta-data pertinent to the hate speech \emph{before} formulating the most appropriate way to respond. Stakeholders considered when the hate speech was created, how often it had been shared and viewed online, asked how many followers the perpetrator has and whether they have a pattern of behaviour in posting such content, and discussed how well the perpetrator seemed to know the target.

Participants from NGOs $A$ and $E$ pointed out that the same hate speech may be shared by a perpetrator with a huge reach online or by a young person in danger of being (further) radicalised, and the strategies they would adopt in those scenarios 
differ. They favoured sarcasm/shaming to respond to someone with a large following, but adopting a kinder/empathetic tone that would encourage someone without such a following to reflect on their behaviour, e.g. responding with \textit{`What if this wasyour sister?'}
NGO $B$ additionally stressed the impor-

\onecolumn
{\fontsize{6.5}{7.6}\selectfont
\begin{longtable}{p{2.5cm}p{2.9cm}p{3.5cm}p{2.6cm}p{2.5cm}}

\textbf{Publication} & \textbf{HS source} & \textbf{CS source} & \textbf{Human input and Task (None = $\times$)} & \textbf{Stakeholder involvement (\checkmark/$\times$) with Details} \\
\hline
\endfirsthead
\caption[]{(continued)} \\
\textbf{Publication} & \textbf{HS source} & \textbf{CS source} & \textbf{Human input, Task} & \textbf{Stakeholder involvement, Details} \\
\hline
\endhead
\citet{Tetzlaff2017DistractinglySexyHS} & N/A & Crawling & Val CS &  $\times$unspecified \\
\citet{Zubiaga2024ALR} & Existing dataset ($\heartsuit$, $\spadesuit$) & Existing dataset ($\heartsuit$, $\spadesuit$) + Automated generation & Ann/Eval CS & $\times$unspecified \\
\citet{Ju2024AMP} & Existing dataset ($\clubsuit$) & Existing dataset ($\clubsuit$) + Automated generation & $\times$ & $\times$no human input \\
\citet{Jones2024AMF} & Existing dataset ($\spadesuit$) & Existing dataset ($\spadesuit$) + Automated generation & Ann/Eval CS & \emph{Possibly}: crowdworkers \\
\citet{Borrelli2022AQA} & Crawling & Crawling & $\times$ & $\times$no human input \\
\citet{Lee2023AlternativeSC} & Existing dataset ($\spadesuit$)& Human annotation & \begin{tabular}[t]{@{}l@{}}Val CS + Ann/Eval CS\end{tabular} & \emph{Possibly}: online 6+ hrs/day \\
\citet{mathew2018analyzing} & Crawling & Crawling & Ann/Eval CS & $\times$unspecified \\
\citet{Song2024AssessingTH} & Crawling & Existing dataset ($\heartsuit$, $\spadesuit$, $\Diamond$) + Crawling & Ann/Eval CS & $\times$academics \\
\citet{Rodrguez2023AutomaticCG} & Existing dataset ($\Box$) & Existing dataset ($\Box$) & Edit MT HS/CS & $\times$academics \\
\citet{Bengoetxea2024BasqueAS} & Existing dataset ($\heartsuit$) & Existing dataset ($\heartsuit$) & \begin{tabular}[t]{@{}l@{}}Edit MT HS/CS + Ann/Eval CS\end{tabular} & $\times$professional and native Spanish+Basque \\
\citet{Ping2024BehindTC} & Existing dataset ($\spadesuit$, other) & Crowdsourcing & \begin{tabular}[t]{@{}l@{}}Write CS + Ann/Eval CS\end{tabular} & \emph{Possibly}: crowdworkers \\
\citet{Mun2023BeyondDH} & Existing dataset ($\spadesuit$, other) + Crawling & Author written + Automated generation & \begin{tabular}[t]{@{}l@{}}Write CS + Ann/Eval CS\end{tabular} & $\times$authors, crowdworkers \\
\citet{Bennie2025CODEOFCONDUCTAM} & Existing dataset ($\spadesuit$) & Automated generation & $\times$ & $\times$no human input \\
\citet{Saha2024ConsolidatingSF} & Existing dataset ($\heartsuit$, $\clubsuit$) & Existing dataset ($\heartsuit$, $\clubsuit$) + Automated generation & Ann/Eval CS & $\times$crowdworkers \\
\citet{Cima2024ContextualizedCS} & Crawling & Existing dataset ($\spadesuit$, $\Diamond$) + Crawling + Automated generation & Ann/Eval CS & $\times$crowdworkers \\
\citet{Santamaria2024ContextualizedGR} & Existing dataset ($\clubsuit$) & Existing dataset ($\clubsuit$) + Automated generation & Ann/Eval CS & $\times$crowdworkers \\
\citet{Garland2023CorrectionIA} & Crawling & Crawling & Val HS/CS &  $\times$authors, crowdworkers \\
\citet{Zhang2024COTAG} & Existing dataset ($\spadesuit$, $\Diamond$) & Existing dataset ($\spadesuit$, $\Diamond$) + Automated generation & Ann/Eval CS & $\times$unspecified \\
\citet{Langer2019CounterNI} & Crawling & Crawling & Qualitative analysis CS & $\times$authors \\
\citet{Saha2022CounterGeDiAC} & Existing dataset ($\heartsuit$, $\Diamond$) & Existing dataset ($\heartsuit$, $\Diamond$) + Automated generation & Ann/Eval CS & $\times$academics \\
\citet{Garland2020CounteringHO} & Crawling & Crawling & Ann/Eval CS & \begin{tabular}[t]{@{}l@{}}$\times$native German crowdworkers\end{tabular} \\
\citet{Ding2024CounterQuillIT} & Existing dataset ($\spadesuit$, other) & Hybrid & Write CS & \emph{Possibly}: crowdworkers \\
\citet{Mun2024CounterspeakersPU} & - & - & Opinons on CS & \begin{tabular}[t]{@{}l@{}}$\checkmark$NGO workers, crowdworkers\end{tabular} \\
\citet{Saha2024CrowdCounterAB} & Existing dataset (other) & Crowdsourcing & \begin{tabular}[t]{@{}l@{}}Write CS +Ann/Eval CS\end{tabular} & \begin{tabular}[t]{@{}l@{}}$\times$crowdworkers, academics\end{tabular} \\
\citet{Hengle2025CSEvalTA} & Existing dataset (other) & Nichesourcing & Ann/Eval CS & $\times$ academics \\
\citet{Hassan2023DisCGenAF} & Hybrid & Hybrid + Automated generation & \begin{tabular}[t]{@{}l@{}}Val HS/CS + Ann/Eval HS/CS\\ + Edit CS\end{tabular} & $\times$academics \\
\citet{Song2025EchoesOD} & Crawling & Crawling & Val CS & $\times$authors \\
\citet{Chung2021EmpoweringNI} & Crawling & Hybrid & \begin{tabular}[t]{@{}l@{}}Edit CS + Ann/Eval CS\end{tabular} & $\checkmark$NGO workers \\
\citet{Wang2024F2RLFA} & Existing dataset ($\heartsuit$, $\spadesuit$, and $\Box$) & Automated generation & $\times$ & $\times$no human input \\
\citet{Zhu2021GeneratePS} & Existing dataset ($\heartsuit$, $\Diamond$) & Automated generation & Ann/Eval CS & $\times$native English \\
\citet{Tekiroglu2020GeneratingCN} & Existing dataset ($\heartsuit$, $\Diamond$, other) & Hybrid & Val CS + Edit CS & $\checkmark$NGO workers \\
\citet{Bar2024GenerativeAM} & Crawling & Crawling &  $\times$ &  $\times$no human input \\
\citet{Yu2022HateSA} & Crawling & Crawling & Ann/Eval HS/CS & $\times$crowdworkers \\
\citet{Alyahya2024HatredSF} & Existing dataset ($\clubsuit$, other) & Existing dataset ($\clubsuit$, other) & Ann/Eval CS & $\times$crowdworkers \\
\citet{Furman2023HighqualityAI} & Existing dataset (other) & Existing dataset (other) & Ann/Eval CS & $\times$authors, academics \\
\citet{Hickey2024HostileCD} & Crawling & Crawling & Ann/Eval CS & $\times$authors, academics \\
\citet{Tonini2024HowDW} & Crawling & Crawling & \begin{tabular}[t]{@{}l@{}}Val HS/CS + Ann/Eval CS\end{tabular} & \begin{tabular}[t]{@{}l@{}}$\checkmark$NGO workers, $\times$academics\end{tabular} \\
\citet{Saha2024IntegratingAA} & Existing dataset ($\heartsuit,\spadesuit,\clubsuit$, other) & Existing dataset ($\heartsuit$, $\spadesuit$, $\clubsuit$, other) & Ann/Eval CS & \begin{tabular}[t]{@{}l@{}}$\times$crowdworkers, academics\end{tabular} \\
\citet{Wang2024IntentAwareAH} & Existing dataset ($\heartsuit$, $\spadesuit$) &  Existing dataset ($\heartsuit$, $\spadesuit$) & Ann/Eval CS &  $\times$unspecified \\
\citet{Hengle2024IntentconditionedAN} & Existing dataset (other) & Existing dataset (other) & Ann/Eval CS &  $\times$academics \\
\citet{Mathew2020InteractionDB} & Crawling & Crawling & Ann/Eval HS/CS & $\times$academics \\
\citet{bonaldi-etal-2024-safer}. &  Existing dataset (other) & Automated generation &  Ann/Eval CS &   $\times$academics \\
\citet{Chung2020ItalianCN} &  Existing dataset ($\heartsuit$) & Existing dataset ($\heartsuit$) &  Ann/Eval CS &  $\times$native Italian \\
\citet{Zubiaga2024IxaAR} &  Existing dataset (other) & Existing dataset (other) & Ann/Eval CS &   $\times$unspecified \\
\citet{Lee2024LeveragingPR} & Existing dataset (other) & Existing dataset (other) &  $\times$ &   $\times$no human input \\
\citet{Das2024LowResourceCG} & Crawling & Crowdsourcing & \begin{tabular}[t]{@{}l@{}}Val HS/CS + Write CS\end{tabular} & $\times$academics \\
\citet{Chung2021MultilingualCN} & Existing dataset ($\heartsuit$) & Existing dataset ($\heartsuit$) & $\times$ & $\times$no human input \\
\citet{gligoric-etal-2024-nlp} & Existing dataset ($\spadesuit$, $\Box$, other) & Existing dataset ($\spadesuit$, $\Box$, other) & Ann/Eval HS/CS & $\times$unspecified \\
\citet{Wadhwa2024NortheasternUA} & Existing dataset ($\spadesuit$) & Existing dataset (other) & $\times$ &  $\times$no human input \\
\citet{Chung2024OnTE} &  Existing dataset (other) + Hybrid &  Hybrid + Automated generation &  \begin{tabular}[t]{@{}l@{}}Write CS + Ann/Eval CS\end{tabular} & \begin{tabular}[t]{@{}l@{}}$\checkmark$NGO workers, $\times$crowdworkers\end{tabular} \\
\citet{Saha2024OnZC} & Existing dataset ($\heartsuit$, $\spadesuit$, $\Diamond$) & Existing dataset ($\heartsuit$, $\spadesuit$, $\Diamond$) & $\times$ &   $\times$no human input \\
\citet{Hong2024OutcomeConstrainedLL} & Existing dataset ($\Diamond$, other) & Existing dataset ($\Diamond$) + Automated generation & Ann/Eval CS & $\times$academics\\
\citet{Rodriguez2024OverviewOR} & Existing dataset ($\spadesuit$) + Crawling? & Existing dataset ($\spadesuit$) + Nichesourcing? & \begin{tabular}[t]{@{}l@{}}Edit MT HS/CS + Write CS \\+ Ann/Eval CS\end{tabular} & $\times$unspecified \\
\citet{Bennie2025PANDAP} &  Existing dataset (other) + Automated generation & Hybrid & \begin{tabular}[t]{@{}l@{}}Ann/Eval CS + Edit CS\end{tabular} & $\times$academics \\
\citet{Furman2022ParsimoniousAA} & Existing dataset (other) & Crowdsourcing? & \begin{tabular}[t]{@{}l@{}}Ann/Eval HS + Write CS\end{tabular} &  $\times$unspecified\\
\citet{Ping2024PerceivingAC} & Existing dataset ($\spadesuit$, other) & Crowdsourcing & \begin{tabular}[t]{@{}l@{}}Val HS + Write CS \\ + Ann/Eval CS\end{tabular} & \begin{tabular}[t]{@{}l@{}}$\checkmark$crowdsourcing + authors\end{tabular} \\
\citet{Ziems2020RacismIA} &  Hybrid + Automated detection & Hybrid + Automated detection & Ann/Eval HS/CS &  $\times$academics \\
\citet{Peng2024RescuingCA} &  Existing dataset (other) &  Existing dataset (other) & Ann/Eval CS &  $\times$unspecified \\
\citet{Jiang2023ReZGRZ} &  Existing dataset ($\spadesuit$) & Existing dataset ($\spadesuit$) & Ann/Eval CS &  $\times$crowdworkers \\
\citet{Saha2023SelfsupervisionAC} & Existing dataset (Unspecified) & Existing dataset (Unspecified) & $\times$ &  $\times$no human input \\
\citet{Arpinar2016SocialMA} & &  &  $\times$ &  $\times$no human input \\
\citet{Alsagheer2023StatisticalAO} &  N/A &  Crawling &  $\times$ &  $\times$no human input \\
\citet{Mathew2018ThouSN} & Crawling & Crawling & Ann/Eval CS &  $\times$academics \\
\citet{Tekiroglu2022UsingPL} & Existing dataset ($\spadesuit$) & Existing dataset ($\spadesuit$) &  Ann/Eval CS &  $\times$unspecified \\
\citet{Leekha2024WarOW} & Hybrid &  Automated generation & Ann/Eval CS & $\times$unspecified \\
\citet{bonaldi-etal-2023-weigh} & Existing dataset ($\spadesuit$) & Existing dataset ($\spadesuit$) & Ann/Eval CS & $\times$unspecified \\
\citet{Halim2023WokeGPTIC} & Hybrid: (uses $\heartsuit$) &  Existing dataset ($\heartsuit$) & Ann/Eval CS &  $\times$academics \\
\caption{Summary of included resources for counterspeech with the same dataset labels and column description from \autoref{tab:source_datasets} (\small Key:  $\heartsuit$ CONAN, $\spadesuit$ Multi-target CONAN, $\clubsuit$ DIALOCONAN, $\Box$ MTKGCONAN and  $\Diamond$ Benchmark)}
\label{tab:datasets}
\end{longtable}}
\twocolumn

\noindent tance of educational responses to counter oGBV in such cases, pointing out the lack of educational content that addresses young men who feel alienated. NGO $A$ suggested having different strategies even \emph{within the response} conditioned on different roles, i.e. shutting down/not engaging the perpetrator.\footnote{and noted that some charities have strict policies against engaging the perpetrator.}, providing support or resources for the target and education for the bystanders.

NGOs $C$ and $D$ discussed trends of oGBV in smaller communities and ethnic minority groups; often the perpetrator knows the target personally and will try to socially isolate them from their community by spreading lies or private information (e.g. images) about them. Thus how well the perpetrator knows the target matters; countering targeted harassment will not be the same as countering online bullying or dogpiling. 

In the community, it is somewhat of a norm to prioritise the metadata of the annotator, i.e. providing demographic information such as age, educational background and gender.\footnote{Demographics have become the norm to provide with paper submissions to ACL, as shown here https://aclrollingreview.org/responsibleNLPresearch/.}
In contrast, the results of our survey show that NLP counterspeech research does not focus attention on metadata related to the hate speech itself, i.e. it is not present in existing counterspeech datasets and in turn affects research that uses existing datasets (nearly $50\%$ of the resources we surveyed).  We additionally find that $\approx 43\%$ of the resources do not even mention the target group of the hate speech, in particular for those resources using existing datasets. Among the resources that do mention the target, most of them do not consider the information in their design, analysis or evaluation.

While some efforts exist to further sub-categorise GBV in hate speech detection (for instance, \emph{benevolent} vs. \emph{hostile} sexism -- see \citet{abercrombie-etal-2023-resources} for an overview), none of the counterspeech resources including the source datasets in \autoref{tab:source_datasets} have such fine-grained categorisation (e.g. harassment vs. dogpiling) -- i.e. it would not be possible to condition counterspeech 
responses specific to the sub-category as discussed by NGOs $C$ and $D$. 
While a recent trend in automated counterspeech generation is to utilise strategies originally proposed by \citet{benesch2016counterspeech}, these methods are limited by the available linguistic cues present in the hate speech, so strategy generation is not holistic, e.g. considering the audience reach of the perpetrator. 
Furthermore, to the best of our knowledge, no information on \emph{who} the counterspeech addresses i.e. perpetrator, bystander or target is present in existing resources. Thus NLP counterspeech resources focus on \emph{what} was said in the hate speech given the lack of other metadata available, whereas stakeholders additionally give importance to the surrounding context. 

\paragraph{Anthropomorphism.} Some interesting issues were raised around the perceived origins of AI-generated counterspeech. Stakeholders from NGO $E$ unanimously agreed that it should be made clear that any counterspeech is artificially generated and not produced by a human. This raises questions of how much store people will put into the responses if they know it is generated by a `bot'.
NGO $A$ discussed being wary of bots reinforcing what are stereotypically considered \emph{`feminazi'} talking points, and that having an anthropomorphically humorous bot is preferable.
In the focus group with NGO $E$, opinion was divided on whether the ‘bot’ delivering the counterspeech should be explicitly gendered, and if so, how this might impact the effectiveness of its message. There was a consensus that a female persona should not be employed, due to the risk of the message being ignored or diminished as a result. Following this logic, some felt that a male persona would have greater credibility with perpetrators, making them more receptive to the counterspeech message. However, this was objected to by others who felt the bot should strive to be gender neutral / ungendered -- although we note this is difficult to achieve, as people still attribute binary gender to systems despite having minimal gender markers \citep{aylet_2019,abercrombie-etal-2023-mirages}.

\paragraph{The temporality of language and `algo-speak'.}
Resources like datasets encode the context of the period in which the data has been collected and annotated. 
NGOs $A$ and $C$ brought up that perpetrators often engage in \emph{`algospeak'}, i.e. finding ways to escape being flagged by content moderation algorithms. However, NGO $A$ also stated that perpetrators on newer social media platforms simply repackage oGBV in newer ways; i.e. the implicit nature remains the same. 

\subsection{Recommendations}\label{sec:recommendations}
In this section, we distil the results of the focus groups into a practical set of data features that are desirable to collect, which could potentially bridge the gap between how counterspeech is tackled in the real world by stakeholders versus counterspeech-focused NLP.

(\textsc{automatically collected}) \textbf{Contextual information}, such as \textbf{meta-data} from social media \cite{10076443} (e.g. the number of followers the perpetrator has, how much the hate speech has been viewed and shared) is needed to determine which strategy to adopt. Further \textbf{dialogue context} will allow for annotators to make better informed decisions \cite{sandri-etal-2023-dont}. While difficult to determine, it may also reveal information about the connection the perpetrator has to the target (e.g. repeated hate speech within the same dialogue).

(\textsc{requiring annotator education}) The \textbf{sub-category of hate}, for instance, if the sub-category of oGBV is dogpiling, counterspeech generation at scale may be required by prioritising quantity over quality. 
The \textbf{roles}; i.e. paying attention to who is involved and the impact: targets, perpetrators and bystanders.  
A consensus is emerging that bystander involvement is the key to change. Bystander intervention \citep{ward_2022} has skyrocketed as a pivotal concept in contemporary GBV studies, where evidence shows that their behavioural decisions, shaped by many socio-cultural and psychological variables \citep{mainwaring2023systematic}, are key to GBV outcomes, such as prevention, reporting, and harm-reduction. 
    
(\textsc{requiring stakeholder input}) \textbf{Instances of illegal language}, i.e. whether the hate speech contains illegal language and \textbf{resources} that educate the bystander and provide support for the target. These may involve working with stakeholders to compile resources on a local level, or consulting stakeholder written sources for up to date factual and educational responses. 

\section{Conclusions}
We systematically reviewed the current state of counterspeech research in NLP.
We found that there has been something of a downturn in the extent to which affected stakeholders are engaged in participatory design for this task, with the field heavily relying on a few key datasets and human input limited to a large extent to computer science researchers. 
To encourage more participatory approaches to NLP counterspeech research, we make recommendations based on feedback from focus groups engaged in tackling real-world hate speech.

\section*{Limitations}
This survey focuses exclusively on peer-reviewed NLP and computational social science publications.
It does not experimentally validate the impact of stakeholder-informed methods on counterspeech effectiveness. Future research direction requires assessing how such methods for counterspeech could influence the real-world outcomes.
Besides, the participatory case study only collaborates with five NGOs with a specific focus on online Gender-Based Violence, which may not fully capture the perspectives of other affected communities, such as religious, or LGBTQ+ groups, etc.

\section*{Ethical Considerations}
This study was approved by our Institutional Review Board (IRB),
of the School of Mathematical and Computer Sciences at Heriot-Watt University
which reviewed our methodologies and protocols to ensure compliance with ethical standards.
Our participatory case study with NGOs was conducted with informed consent, and all participants were made aware of the goals of the research, how their input would be used, and their right to withdraw at any time. 
Given the sensitive nature of online Gender-Based Violence, we anonymised all identifying details of participants from NGOs, but will release the organisations' names upon acceptance. Furthermore, we compensated the NGOs fairly for their time spent in our focus groups, discussing within our network what is a standard rate for their expertise.\looseness=-1   

\bibliography{anthology,custom}

\appendix
\section{Resource publication years}
\begin{figure}[ht!]
    \centering
    \includegraphics[width=1.1\columnwidth]{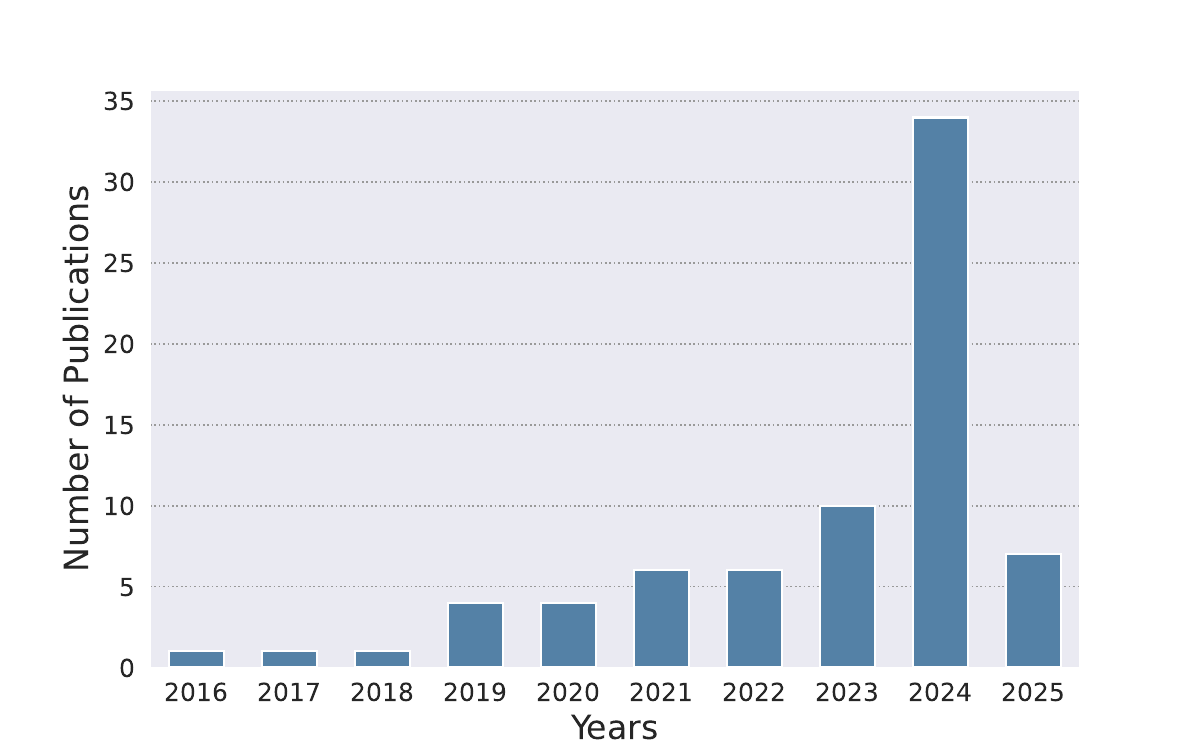}
    \caption{Publications per year up to March 2025.}
    \label{fig:publication}
\end{figure}

\autoref{fig:publication} shows the resources we surveyed by publication year, with a notable recent spike. 

\section{Full table of resources for counterspeech}
\autoref{tab:datasets_appendix} shows all the resources we considered in our survey using the labels and column description from \autoref{tab:source_datasets}.

\label{sec:appendix}

\onecolumn
{\fontsize{6.5}{7.6}\selectfont
\begin{longtable}{p{2.6cm}p{3.5cm}p{3.5cm}p{2.3cm}p{2.4cm}}

\textbf{Publication} & \textbf{HS source} & \textbf{CS source} & \textbf{Human input and Task (None = $\times$)} & \textbf{Stakeholder involvement (\checkmark/$\times$) with Details} \\

\hline
\endfirsthead
\caption[]{(continued)} \\
\textbf{Publication} & \textbf{HS source} & \textbf{CS source} & \textbf{Human input and Task (None = $\times$)} & \textbf{Stakeholder involvement (\checkmark/$\times$) with Details} \\
\hline
\endhead
\citet{Tetzlaff2017DistractinglySexyHS} & N/A & Crawling & Validate CS &  $\times$unspecified \\
\citet{Zubiaga2024ALR} & Existing dataset ($\heartsuit$, $\spadesuit$) & Existing dataset ($\heartsuit$, $\spadesuit$) + Automated generation & Annotate/Evaluate CS & $\times$unspecified \\
\citet{Ju2024AMP} & Existing dataset ($\clubsuit$) & Existing dataset ($\clubsuit$) + Automated generation & $\times$ & $\times$no human input \\
\citet{Jones2024AMF} & Existing dataset ($\spadesuit$) & Existing dataset ($\spadesuit$) + Automated generation & Annotate/Evaluate CS & \emph{Possibly}: crowdworkers \\
\citet{Borrelli2022AQA} & Crawling & Crawling & $\times$ & $\times$no human input \\
\citet{Lee2023AlternativeSC} & Existing dataset ($\spadesuit$)& Human annotation & \begin{tabular}[t]{@{}l@{}}Validate CS + \\ Annotate/Evaluate CS\end{tabular} & \emph{Possibly}: people online 6+ hrs/day \\
\citet{mathew2018analyzing} & Crawling & Crawling & Annotate/Evaluate CS & $\times$unspecified \\
\citet{Song2024AssessingTH} & Crawling & Existing dataset ($\heartsuit$, $\spadesuit$, $\Diamond$) + Crawling & Annotate/Evaluate CS & $\times$academics \\
\citet{Rodrguez2023AutomaticCG} & Existing dataset ($\Box$) & Existing dataset ($\Box$) & Edit MT HS/CS & $\times$academics \\
\citet{Bengoetxea2024BasqueAS} & Existing dataset ($\heartsuit$) & Existing dataset ($\heartsuit$) & \begin{tabular}[t]{@{}l@{}}Edit MT HS/CS +\\ Annotate/Evaluate CS\end{tabular} & $\times$professional translators Spanish+Basque, native Spanish+Basque annotators \\
\citet{Ping2024BehindTC} & Existing dataset (ETHOS \citep{mollas2022ethos}, $\spadesuit$, Multilingual and Multi-Aspect Hate Speech Analysis \citep{ousidhoum-etal-2019-multilingual}) & Crowdsourcing & \begin{tabular}[t]{@{}l@{}}Write CS + \\ Annotate/Evaluate CS\end{tabular} & \emph{Possibly}: crowdworkers \\
\citet{Mun2023BeyondDH} & Existing dataset (Social Bias Inference corpus \citep{sap-etal-2020-social},  $\spadesuit$,  Winning Argument Corpus \citep{tan2016winning}) + Crawling & Author written + Automated generation & \begin{tabular}[t]{@{}l@{}}Write CS + \\ Annotate/Evaluate CS\end{tabular} & $\times$authors, crowdworkers \\
\citet{Bennie2025CODEOFCONDUCTAM} & Existing dataset ($\spadesuit$) & Existing dataset (ML-MTCONAN-KN \citep{bonaldi-etal-2025-first}) & $\times$ & $\times$no human input \\
\citet{chung-etal-2019-conan} & Nichesourcing & Nichesourcing & \begin{tabular}[t]{@{}l@{}}Write HS/CS +\\ Paraphrase CS\end{tabular} & \begin{tabular}[t]{@{}l@{}}$\checkmark$NGO workers, $\times$non-experts\end{tabular} \\
\citet{Saha2024ConsolidatingSF} & Existing dataset ($\heartsuit$, $\clubsuit$) & Existing dataset ($\heartsuit$, $\clubsuit$) + Automated generation & Annotate/Evaluate CS & $\times$crowdworkers \\
\citet{Cima2024ContextualizedCS} & Crawling & Existing dataset ($\spadesuit$, $\Diamond$) + Crawling + Automated generation & Annotate/Evaluate CS & $\times$crowdworkers \\
\citet{Santamaria2024ContextualizedGR} & Existing dataset ($\clubsuit$) & Existing dataset ($\clubsuit$) + Automated generation & Annotate/Evaluate CS & $\times$crowdworkers \\
\citet{Garland2023CorrectionIA} & Crawling & Crawling & Validate HS/CS &  $\times$authors, crowdworkers \\
\citet{Zhang2024COTAG} & Existing dataset ($\spadesuit$, $\Diamond$) & Existing dataset ($\spadesuit$, $\Diamond$) + Automated generation & Annotate/Evaluate CS & $\times$unspecified \\
\citet{Langer2019CounterNI} & Crawling & Crawling & Qualitative analysis CS & $\times$authors \\
\citet{Saha2022CounterGeDiAC} & Existing dataset ($\heartsuit$, $\Diamond$) & Existing dataset ($\heartsuit$, $\Diamond$) + Automated generation & Annotate/Evaluate CS & $\times$academics \\
\citet{Garland2020CounteringHO} & Crawling & Crawling & Annotate/Evaluate CS & $\times$native German crowdworkers \\
\citet{Ding2024CounterQuillIT} & Existing dataset (ETHOS \citep{mollas2022ethos}, $\spadesuit$, Multilingual and Multi-Aspect Hate Speech Analysis \citep{ousidhoum-etal-2019-multilingual}) & Hybrid: Automated generation and Crowdsourcing & Write CS & \emph{Possibly}: crowdworkers \\
\citet{Mun2024CounterspeakersPU} & & & Opinons on CS & \begin{tabular}[t]{@{}l@{}}$\checkmark$NGO workers +\\ crowdworkers\end{tabular} \\
\citet{gupta-etal-2023-counterspeeches} & Existing dataset ($\spadesuit$) & Existing dataset ($\spadesuit$) + Human written & Write CS & $\times$academics \\
\citet{Saha2024CrowdCounterAB} & Existing dataset (HateXplain \citep{mathew2021hatexplain}) & Crowdsourcing & \begin{tabular}[t]{@{}l@{}}Write CS +\\ Annotate/Evaluate CS\end{tabular} & $\times$crowdworkers, academics \\
\citet{Hengle2025CSEvalTA} & Existing dataset (IntentCONAN \citep{gupta-etal-2023-counterspeeches}) & Nichesourcing & Annotate/Evaluate CS & $\times$academics \\
\citet{Hassan2023DisCGenAF} & Hybrid: Crawling and Automated detection and Human annotation & Hybrid: Crawling and Automated detection and Human annotation + Automated generation & \begin{tabular}[t]{@{}l@{}}Validate HS/CS + \\ Annotate/Evaluate HS/CS + \\ Edit CS\end{tabular} & $\times$academics \\
\citet{Song2025EchoesOD} & Crawling & Crawling & Validate CS & $\times$authors \\
\citet{Chung2021EmpoweringNI} & Crawling & Hybrid: Automated generation and Nichesourcing & \begin{tabular}[t]{@{}l@{}}Edit CS + \\ Annotate/Evaluate CS\end{tabular} & $\checkmark$NGO workers \\
\citet{Wang2024F2RLFA} & Existing dataset ($\heartsuit$, $\spadesuit$, and $\Box$) & Automated generation & $\times$ & $\times$no human input \\
\citet{Zhu2021GeneratePS} & Existing dataset ($\heartsuit$, $\Diamond$) & Automated generation & Annotate/Evaluate CS & $\times$native English \\
\citet{Tekiroglu2020GeneratingCN} & Existing dataset (Twitter dataset \citep{mathew2018analyzing}, $\heartsuit$, $\Diamond$) & Hybrid: Crowdsourcing and Nichesourcing & Validate CS + Edit CS & $\checkmark$NGO workers \\
\citet{Bar2024GenerativeAM} & Crawling & Crawling &  $\times$ &  $\times$no human input \\
\citet{Yu2022HateSA} & Crawling & Crawling & Annotate/Evaluate HS/CS & $\times$crowdworkers \\
\citet{Alyahya2024HatredSF} & Existing dataset ($\clubsuit$, ContextCounter \citep{albanyan-etal-2023-counterhate}) & Existing dataset ($\clubsuit$, ContextCounter \citep{albanyan-etal-2023-counterhate}) & Annotate/Evaluate CS & $\times$crowdworkers \\
\citet{Furman2023HighqualityAI} & Existing dataset (ASOHMO \citep{furman-etal-2023-argumentative}, CONEAS\footnote{\label{coneas}\url{https://github.com/ConeasDataset/CONEAS/}}) & Existing dataset (ASOHMO \citep{furman-etal-2023-argumentative}, CONEAS & Annotate/Evaluate CS & $\times$authors, academics \\
\citet{Hickey2024HostileCD} & Crawling & Crawling & Annotate/Evaluate CS & $\times$authors, academics \\
\citet{Tonini2024HowDW} & Crawling & Crawling & \begin{tabular}[t]{@{}l@{}}Validate HS/CS + \\ Annotate/Evaluate CS\end{tabular} & \begin{tabular}[t]{@{}l@{}}$\checkmark$NGO workers, \\ $\times$academics\end{tabular} \\
\citet{Fanton2021HumanintheLoopFD} & Hybrid: Nichesourcing and Automated (Human-in-the-loop) & Hybrid: Nichesourcing and Automated (Human-in-the-loop) & \begin{tabular}[t]{@{}l@{}}Validate CS + \\ Edit CS\end{tabular} & \begin{tabular}[t]{@{}l@{}}$\checkmark$NGO workers\\ $\times$academics\end{tabular} \\
\citet{bonaldi-etal-2022-human} & Hybrid: Nichesourcing and Automated (Human-in-the-loop) & Hybrid: Nichesourcing and Automated (Human-in-the-loop) & \begin{tabular}[t]{@{}l@{}}Validate CS + \\ Edit CS\end{tabular} & $\checkmark$NGO workers \\
\citet{Saha2024IntegratingAA} & Existing dataset ($\heartsuit$, $\spadesuit$, $\clubsuit$, OUMdials \citep{farag-etal-2022-opening}, MisinfoCorrect \citep{he2023reinforcement}, ASFoCoNG \citep{furman2022parsimonious}) & Existing dataset ($\heartsuit$, $\spadesuit$, $\clubsuit$, TSNH \citep{Mathew2018ThouSN}, ASFoCoNG \citep{furman2022parsimonious}) & Annotate/Evaluate CS & $\times$crowdworkers, academics \\
\citet{Wang2024IntentAwareAH} & Existing dataset ($\heartsuit$, $\spadesuit$) &  Existing dataset ($\heartsuit$, $\spadesuit$) & Annotate/Evaluate CS &  $\times$unspecified \\
\citet{Hengle2024IntentconditionedAN} & Existing dataset (IntentCONAN \citep{gupta-etal-2023-counterspeeches}) & Existing dataset (IntentCONAN \citep{gupta-etal-2023-counterspeeches}) & Annotate/Evaluate CS &  $\times$academics \\
\citet{Mathew2020InteractionDB} & Crawling & Crawling & Annotate/Evaluate HS/CS & $\times$academics \\
\citet{bonaldi-etal-2024-safer}. &  Existing dataset (White Supremacy Forum \citep{de-gibert-etal-2018-hate}) & Automated generation &  Annotate/Evaluate CS &   $\times$academics \\
\citet{Chung2020ItalianCN} &  Existing dataset ($\heartsuit$) & Existing dataset ($\heartsuit$) &  Annotate/Evaluate CS &  $\times$native Italian \\
\citet{Zubiaga2024IxaAR} &  Existing dataset (CONAN-MT-SP \citep{vallecillo-rodriguez-etal-2024-conan}) & Existing dataset (CONAN-MT-SP \citep{vallecillo-rodriguez-etal-2024-conan}) & Annotate/Evaluate CS &   $\times$unspecified \\
\citet{Lee2024LeveragingPR} & Existing dataset ($\spadesuit$, Unsmile \citep{kang2022korean}, APEACH \citep{yang-etal-2022-apeach}, BEEP \citep{moon-etal-2020-beep}, KOLD \citep{jeong-etal-2022-kold}) & Existing dataset ($\spadesuit$) &  $\times$ &   $\times$no human input \\
\citet{Das2024LowResourceCG} & Crawling & Crowdsourcing & \begin{tabular}[t]{@{}l@{}}Validate HS/CS + \\ Write CS\end{tabular} & $\times$academics \\
\citet{Chung2021MultilingualCN} & Existing dataset ($\heartsuit$) & Existing dataset ($\heartsuit$) & $\times$ & $\times$no human input \\
\citet{gligoric-etal-2024-nlp} & Existing dataset ($\spadesuit$, $\Box$, MisinfoCorrect \citep{he2023reinforcement}) & Existing dataset ($\spadesuit$, $\Box$, MisinfoCorrect \citep{he2023reinforcement}) & Annotate/Evaluate HS/CS & $\times$unspecified \\
\citet{Wadhwa2024NortheasternUA} & Existing dataset ($\spadesuit$) & Existing dataset (ML-MTCONAN-KN \citep{bonaldi-etal-2025-first}) & $\times$ &  $\times$no human input \\
\citet{Chung2024OnTE} &  Existing dataset (TOXIGEN \citep{chung-bright-2024-effectiveness}) + Hybrid: Crawling and Human annotation &  Hybrid: Crawling and Nichesourcing + Automated generation &  \begin{tabular}[t]{@{}l@{}}Write CS + \\ Annotate/Evaluate CS\end{tabular} & $\checkmark$civil society org workers, $\times$crowdworkers \\
\citet{Saha2024OnZC} & Existing dataset ($\heartsuit$, $\spadesuit$, $\Diamond$) & Existing dataset ($\heartsuit$, $\spadesuit$, $\Diamond$) & $\times$ &   $\times$no human input \\
\citet{Hong2024OutcomeConstrainedLL} & Existing dataset (CAD \citep{vidgen-etal-2021-introducing}, $\Diamond$) & Existing dataset ($\Diamond$) + Automated generation & Annotate/Evaluate CS & $\times$academics \\
\citet{Rodriguez2024OverviewOR} & Existing dataset ($\spadesuit$) + Crawling? & Existing dataset ($\spadesuit$) + Nichesourcing? & \begin{tabular}[t]{@{}l@{}}Edit MT HS/CS + \\ Write CS + \\ Annotate/Evaluate CS\end{tabular} & $\times$unspecified \\
\citet{Bennie2025PANDAP} &  Existing dataset (COLD \citep{deng-etal-2022-cold}, SWSR \citep{jiang2022swsr}, CHSD \citep{rao-etal-2023-ji}) + Automated generation & Hybrid: Crowdsourcing and Automated generation & \begin{tabular}[t]{@{}l@{}}Annotate/Evaluate CS \\ + Edit CS\end{tabular} & $\times$academics \\
\citet{Furman2022ParsimoniousAA} & Existing dataset (HatEval 2019 \citep{basile-etal-2019-semeval}) & Crowdsourcing? & \begin{tabular}[t]{@{}l@{}}Annotate/Evaluate HS + \\ Write CS\end{tabular} &  $\times$unspecified \\
\citet{Ping2024PerceivingAC} & Existing dataset (ETHOS \citep{mollas2022ethos}, $\spadesuit$, Multilingual and Multi-Aspect Hate Speech Analysis \citep{ousidhoum-etal-2019-multilingual}) & Crowdsourcing & \begin{tabular}[t]{@{}l@{}}Validate HS + \\ Write CS + \\ Annotate/Evaluate CS\end{tabular} & $\checkmark$crowdsourcing,authors \\
\citet{Ziems2020RacismIA} &  Hybrid: Crawling and Human Annotation + Automated detection & Hybrid: Crawling and Human Annotation + Automated detection & Annotate/Evaluate HS/CS &  $\times$academics \\
\citet{Peng2024RescuingCA} &  Existing dataset (Community Notes \citep{wojcik2022birdwatch}) &  Existing dataset (Community Notes \citep{wojcik2022birdwatch}) & Annotate/Evaluate CS &  $\times$unspecified \\
\citet{Jiang2023ReZGRZ} &  Existing dataset ($\spadesuit$) & Existing dataset ($\spadesuit$) & Annotate/Evaluate CS &  $\times$crowdworkers \\
\citet{Saha2023SelfsupervisionAC} & Existing dataset (Unspecified) & Existing dataset (Unspecified) & $\times$ &  $\times$no human input \\
\citet{Arpinar2016SocialMA} & &  &  $\times$ &  $\times$no human input \\
\citet{Alsagheer2023StatisticalAO} &  N/A &  Crawling &  $\times$ &  $\times$no human input \\
\citet{Mathew2018ThouSN} & Crawling & Crawling & Annotate/Evaluate CS &  $\times$academics \\
\citet{Chung2021TowardsKC} & Existing dataset ($\heartsuit$) & Automated generation & \begin{tabular}[t]{@{}l@{}} Annotate/Evaluate CS\end{tabular} &  $\checkmark$NGO workers \\
\citet{Tekiroglu2022UsingPL} & Existing dataset ($\spadesuit$) & Existing dataset ($\spadesuit$) &  Annotate/Evaluate CS &  $\times$unspecified \\
\citet{Leekha2024WarOW} & Hybrid: Crawling and Automated detection &  Automated generation & Annotate/Evaluate CS & $\times$unspecified \\
\citet{bonaldi-etal-2023-weigh} & Existing dataset ($\spadesuit$) & Existing dataset ($\spadesuit$) & Annotate/Evaluate CS & $\times$unspecified \\
\citet{Halim2023WokeGPTIC} & Hybrid: Existing dataset ($\heartsuit$, HateXplain \citep{mathew2021hatexplain}, White Supremacy Forum \citep{de-gibert-etal-2018-hate}, Phoenix Real-Time (PRT) \citep{salam2018distributed}, Expert Domain Corpora, Mainstream Media, Gigaword) and Filtering &  Existing dataset ($\heartsuit$) & Annotate/Evaluate CS &  $\times$academics \\
\citet{qian-etal-2019-benchmark} & Crawling + Crowdsourcing & Crowdsourcing & Validate HS + Write CS & $\times$crowdworkers\\
\citet{vallecillo-rodriguez-etal-2024-conan} & Existing dataset ($\spadesuit$) & Human written & Write CS + Edit MT CS & $\times$academics, translators Spanish, Basque, Italian\\
\hline
\label{tab:datasets_appendix} \\
\caption{Summary of included resources for counterspeech with the same dataset labels and column description from \autoref{tab:source_datasets}. Note: we include datasets from \autoref{tab:source_datasets} and \autoref{tab:datasets}. (KEY: $\heartsuit$ CONAN, $\spadesuit$ Multi-target CONAN, $\clubsuit$ DIALOCONAN, $\Box$ MTKGCONAN and  $\Diamond$ Benchmark).}
\end{longtable}}
\twocolumn

\end{document}